\documentclass{article}
\usepackage{ijcai16}

\usepackage{times}

\usepackage{helvet}
\usepackage{courier}
\usepackage{amssymb}
\usepackage{amsmath}
\usepackage{amsfonts}
\usepackage{graphicx}
\usepackage{sistyle}
\usepackage{bm}
\title{Match-SRNN: Modeling the Recursive Matching Structure with Spatial RNN}
\author{Shengxian Wan$^{*}$, Yanyan Lan$^\dag$, Jun Xu$^\dag$, Jiafeng Guo$^\dag$, Liang Pang$^{*}$, \and Xueqi Cheng$^\dag$\\
CAS Key Lab of Network Data Science and Technology\\
Institute of Computing Technology, Chinese Academy of Sciences, China\\
$^*$\{wanshengxian, pangliang\}@software.ict.ac.cn, $^\dag$\{lanyanyan, junxu, guojiafeng, cxq\}@ict.ac.cn \\
}

\begin{document}
\maketitle
\begin{abstract}
Semantic matching, which aims to determine the matching degree between two texts, is a fundamental problem for many NLP applications. Recently, deep learning approach has been applied to this problem and significant improvements have been achieved. In this paper, we propose to view the generation of the global interaction between two texts as a recursive process: i.e.~the interaction of two texts at each position is a composition of the interactions between their prefixes as well as the word level interaction at the current position. Based on this idea, we propose a novel deep architecture, namely Match-SRNN, to model the recursive matching structure. Firstly, a tensor is constructed to capture the word level interactions. Then a spatial RNN is applied to integrate the local interactions recursively, with importance determined by four types of gates. Finally, the matching score is calculated based on the global interaction. We show that, after degenerated to the exact matching scenario, Match-SRNN can approximate the dynamic programming process of longest common subsequence. Thus, there exists a clear interpretation for Match-SRNN. Our experiments on two semantic matching tasks showed the effectiveness of Match-SRNN, and its ability of visualizing the learned matching structure.
\end{abstract}

\section{Introduction}
Semantic matching is a critical task for many applications in natural language processing, including information retrieval, question answering and paraphrase identification \cite{INR-035}. The target of semantic matching is to determine a matching score for two given texts.
Taking the task of question answering as an example, given a pair of question and answer, a matching function is created to determine the matching degree between these two texts. Traditional methods such as BM25 and feature based learning models usually rely on exact matching patterns to determine the degree, and thus suffer from the vocabulary mismatching problem \cite{INR-035}.

Recently, deep learning approach has been applied to this area and well tackled the vocabulary mismatching problem. Some existing work focus on representing each text as one or several dense vectors, and then calculate the matching score based on the similarity between these vectors. Examples include RAE \cite{socher2011dynamic}, DSSM~\cite{huang2013learning}, CDSSM~\cite{export:226585}, ARC-I~\cite{DBLP:conf/nips/HuLLC14}, CNTN~\cite{DBLP:conf/ijcai/QiuH15},  LSTM-RNN~\cite{palangi2015deep}, MultiGranCNN~\cite{DBLP:conf/naacl/YinS15,DBLP:conf/acl/YinS15} and MV-LSTM~\cite{wan2016aaai}. However, it is usually difficult for these methods to model the complicated interaction relationship between two texts \cite{DBLP:conf/nips/LuL13} because the representations are calculated independently. To address the problem, some other deep methods have been proposed to directly learn the interaction relationship between the two texts, including DeepMatch~\cite{DBLP:conf/nips/LuL13}, ARC-II~\cite{DBLP:conf/nips/HuLLC14}, and MatchPyramid~\cite{pang2016aaai} etc. All these models conducts the matching through a hierarchical matching structure: the global interaction between two texts is a composition of different levels of the local interactions, such as word level and phrase level interactions.

In all of these methods, the mechanism on the generation of the complicated interaction relationship between two texts is not clear, and thus lack of interpretability. In this paper, we propose to tackle the problem in a recursive manner. Specifically, we view the generation of the global interactions as a recursive process. Given two texts $S_1\!\!=\!\!\{w_1,w_2,\cdots,w_m\}$ and $S_2\!\!=\!\!\{v_1,v_2,\cdots,v_n\}$, the interaction at each position $(i,j)$ (i.e.~interaction between $S_1[1{:}i]$ and $S_2[1{:}j]$) is a composition of the interactions between their prefixes (i.e.~three interactions, $S_1[1{:}i{-}1]\!{\sim}\! S_2[1{:}j]$, $S_1[1{:}i]\!{\sim}\! S_2[1{:}j{-}1]$, $S_1[1{:}i{-}1]{\sim} S_2[1{:}j{-}1]$), and the word level interaction at this position (i.e.~the interaction between $w_i$ and $v_j$), where $S[1{:}c]$ stands for the prefix consisting of the previous $c$ words of text $S$. Compared with previous hierarchical matching structure, the recursive matching structure can not only capture the interactions between nearby words, but also take the long distant interactions into account.

Based on the above idea, we propose a novel deep architecture, namely Match-SRNN, to model the recursive matching structure. Firstly, a similarity tensor is constructed to capture the word level interactions between two texts, where each element $\vec{s}_{ij}$ stands for a similarity vector between two words from different texts. Then a spatial (2D) recurrent neural network (spatial RNN) with gated recurrent units is applied to the tensor. Specifically, the representation at each position $\vec{h}_{ij}$ can be viewed as the interactions between the two prefixes, i.e.~${S_1[1{:}i]}$ and ${S_2[1{:}j]}$. It is determined by four factors: $\vec{h}_{i-1,j},\vec{h}_{i,j-1},\vec{h}_{i-1,j-1}$ and the input word level interaction $\vec{s}_{ij}$, depending on the corresponding gates, $z_t,z_l,z_d$, and $z_i$, respectively. Finally, the matching score is produced by a linear scoring function on the representation of the global interaction $\vec{h}_{mn}$, obtained by the aforementioned spatial RNN.

We show that Match-SRNN can well approximate the dynamic programming process of longest common subsequence (LCS) problem~\cite{WikiLCS}. Furthermore, our simulation experiments show that a clear matching path can be obtained by backtracking the maximum gates at each position, similar to that in LCS.  Thus, there is a clear interpretation on how the global interaction is generated in Match-SRNN.

We conducted experiments on question answering and paper citation tasks to evaluate the effectiveness of our model.
The experimental results showed that Match-SRNN can significantly outperform existing deep models. Moreover, to visualize the learned matching structure, we showed the matching path of two texts sampled from the real data.

The contributions of this paper can be summarized as:
\begin{itemize}
\item The idea of modeling the mechanism of semantic matching recursively, i.e.~the recursive matching structure.
\item The proposal of a new deep architecture, namely Match-SRNN, to model the recursive matching structure. Experimental results showed that Match-SRNN can significantly improve the performances of semantic matching, compared with existing deep models.
\item The reveal of the relationship between Match-SRNN and the LCS, i.e.~Match-SRNN can reproduce the matching path of LCS in an exact matching scenario.
\end{itemize}
\section{Related Work}
Existing deep learning methods for semantic matching can be categorized into two groups.

One paradigm focuses on representing each text to a dense vector, and then compute the matching score based on the similarity between these two vectors. For example, DSSM~\cite{huang2013learning} uses a multi-layer fully connected neural network to encode a query (or a document) as a vector.
CDSSM~\cite{export:226585} and ARC-I~\cite{DBLP:conf/nips/HuLLC14} utilize convolutional neural network (CNN), while LSTM-RNN~\cite{palangi2015deep} adopts recurrent neural network with long short term memory (LSTM) units to better represent a sentence.
Different from above work, CNTN~\cite{DBLP:conf/ijcai/QiuH15} uses a neural tensor network to model the interaction between two sentences instead of using the cosine function. With this way, it can capture more complex matching relations. Some methods even try to match two sentences with multiple representations, such as words, phrases, and sentences level representations. Examples include RAE~\cite{socher2011dynamic}, BiCNN~\cite{DBLP:conf/naacl/YinS15}, MultiGranCNN~\cite{DBLP:conf/acl/YinS15}, and MV-LSTM~\cite{wan2016aaai}. In general, the idea behind the approach is consistent with users' experience that the matching degree between two sentences can be determined once the meanings of them being well captured. However, it is usually difficult for these methods to model the complicated interaction relationship between two texts, especially when they have already been represented as a compact vector~\cite{DBLP:conf/nips/LuL13,bahdanau2014neural}.

The other paradigm turns to directly model the interaction relationship of two texts. Specifically, the interaction is represented as a dense vector, and then the matching score can be produced by integrating such interaction. Most existing work of this paradigm create a hierarchical matching structure, i.e.~the global interaction between two texts is generated by compositing the local interactions hierarchically.
For example, DeepMatch~\cite{DBLP:conf/nips/LuL13} models the generation of the global interaction between two texts as integrating local interactions based on hierarchies of the topics. MatchPyramid~\cite{pang2016aaai} uses a CNN to model the generation of the global interaction as an abstraction of the word level and phrase level interactions. Defining the matching structure hierarchically has limitations, since hierarchical matching structure usually relies on a fixed window size for composition, the long distant dependency between the local interactions cannot be well captured in this kind of models.

\section{The Recursive Matching Structure}
In all existing methods, the mechanism of semantic matching is complicated and hard to interpret.
In mathematics and computer science, when facing a complicated object, a common method of simplification is to divide a problem into subproblems of the same type, and try to solve the problems recursively. This is the well-known thinking of recursion. In this paper, we propose to tackle the semantic matching problem recursively. The recursive rule is defined as follows.
\newtheorem{definition}{Definition}
\begin{definition}[Recursive Matching Structure]
Given two texts ${S_1{=}\{w_1,\cdots,w_m\}}$ and $S_2{=}{\{v_1,\cdots,v_n\}}$, the interaction between prefixes $S_1[1{:}i]{=}\{w_1,\cdots,w_i\}$ and $S_2[1{:}j]{=}\{v_1,\cdots,v_j\}$ (denoted as $\vec{h}_{ij}$) is composited by the interactions between the sub-prefixes as well as the word level interaction of the current position, as shown by the following equation:
\begin{equation}\label{eq:RecursiveMatchStruct}
\vec{h}_{ij}=f(\vec{h}_{i-1,j},\vec{h}_{i,j-1},\vec{h}_{i-1,j-1},\vec{s}(w_i,v_j)),
\end{equation}
where $\vec{s}(w_i,v_j)$ stands for the interaction between words $w_i$ and $v_j$.
\end{definition}

\begin{figure}[t]
 \centering
 \includegraphics[width=0.33\textwidth]{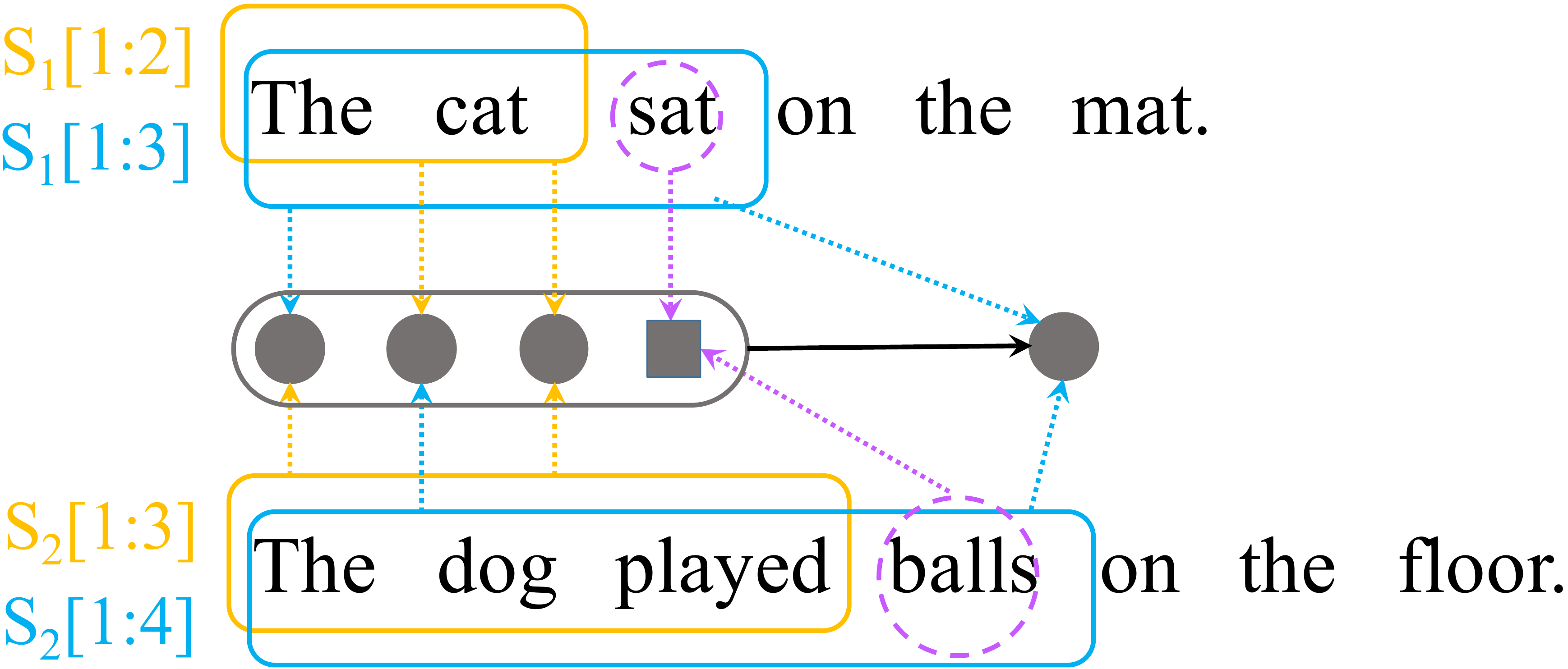}
 \caption{Illustration of the recursive matching structure.}
 \label{fig:example}
 \end{figure}

Figure~\ref{fig:example} illustrates an example of the recursive matching structure for sentences ${S_1{=}\{\text{\em The cat sat on the mat}\}}$ and ${S_2{=}\{\text{\em The dog played balls on the floor}\}}$. Considering the interaction between ${S_1[1{:}3]{=}\{\text{\em The cat sat}\}}$ and ${S_2[1{:}4]{=}\{\text{\em The dog played balls}\}}$ (i.e.~$\vec{h}_{34}$), the recursive matching structure defined above indicates that it is the composition of the interactions between their prefixes (i.e.~$\vec{h}_{24}$, $\vec{h}_{33}$, and $\vec{h}_{23}$) and the word level interaction between `{\em sat}' and `{\em balls}', where $\vec{h}_{24}$ stands for the interaction between $S_1[1{:}2]{=}\{\text{\em The cat}\}$ and $S_2[1{:}4]{=}\{\text{\em The dog played balls}\}$, $\vec{h}_{33}$ denotes the interaction between ${S_1[1{:}3]{=}\{\text{\em The cat sat}\}}$ and ${S_2[1{:}3]{=}\{\text{\em The dog played}\}}$, and $\vec{h}_{23}$ denotes the interaction between ${S_1[1{:}2]{=}\{\text{\em The cat}\}}$ and ${S_2[1{:}3]{=}\{\text{\em The dog played}\}}$.
We can see that the most important interaction, i.e. the interaction between $S_1[1{:}3]{=}\{\text{\em The cat sat}\}$ and $S_2[1{:}3]{=}\{\text{\em The dog played}\}$, has been utilized for representing $\vec{h}_{34}$, which consists well with the human understanding.
Therefore, it is expected that this recursive matching structure can well capture the complicated interaction relationship between two texts because all of the interactions between prefixes have been taken into consideration.
Compared with the hierarchical one, the recursive matching structure is able to capture long-distant dependency among interactions.

\section{Match-SRNN}
\begin{figure}[t]
 \centering
 \includegraphics[width=0.45\textwidth]{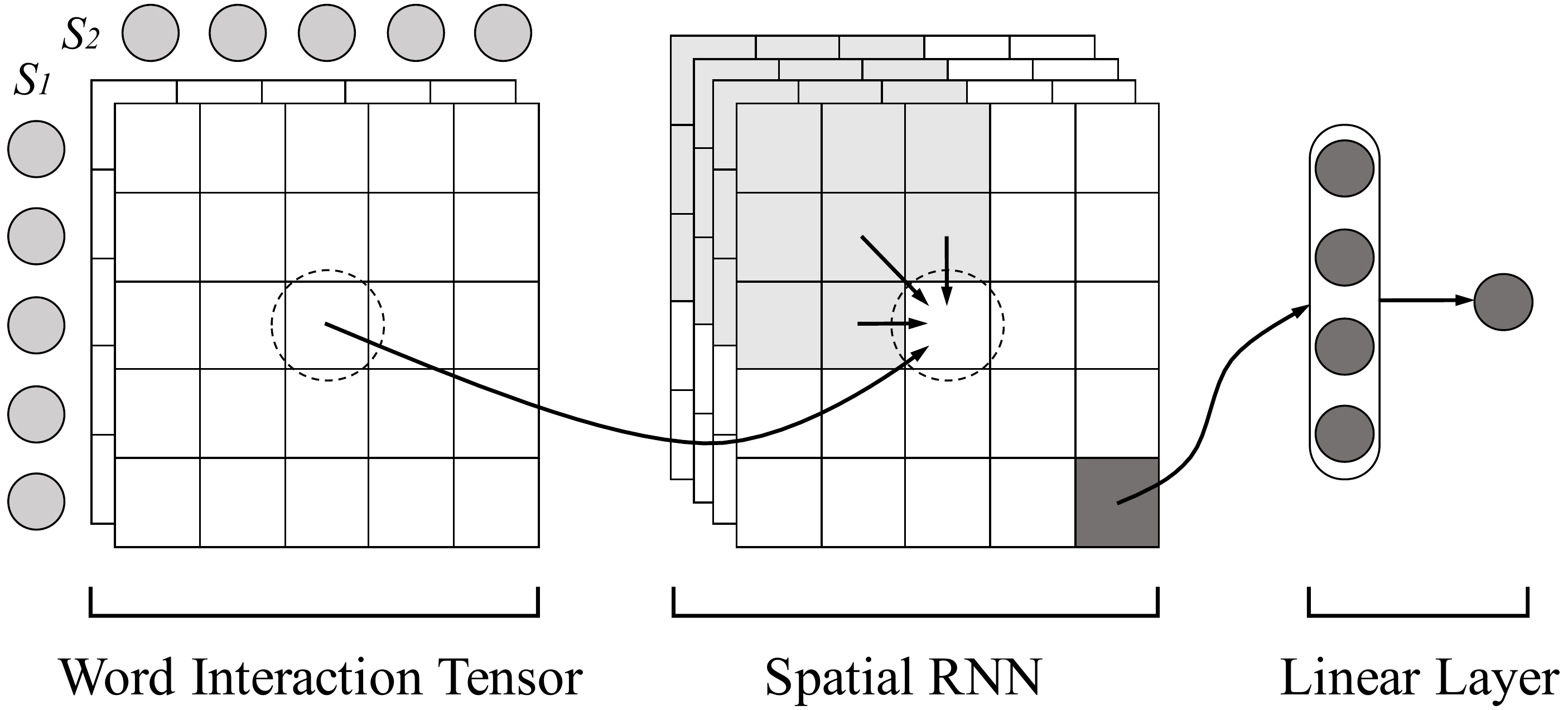}
 \caption{The architecture of Match-SRNN.}
 \label{fig:architecture}
 \end{figure}

In this section, we introduce a new deep architecture, namely Match-SRNN, to model the recursive matching structure. As shown in Figure~\ref{fig:architecture}, Match-SRNN consists of three components:
(1) a neural tensor network to capture the word level interactions; (2) a spatial RNN applied on the word interaction tensor to obtain the global interaction; (3) a linear scoring function to obtain the final matching score.
\subsection{Neural Tensor Network}
In Match-SRNN, a neural tensor network is first utilized to capture the basic interactions between two texts, i.e.~word level interactions.
Specifically, each word is first represented as a distributed vector. Given any two words $w_i$ and $v_j$, and their vectors $u(w_i)$ and $u(v_j)$, the interaction between them can be represented as a vector:
\begin{equation*}
\vec{s}_{ij}=F(u(w_i)^TT^{[1:c]}u(v_j)+W\begin{bmatrix}
		u(w_i)\\
		u(v_j)
	\end{bmatrix}+\vec{b}),
\end{equation*}
where $T^i, i\in[1,...,c]$ is one slice of the tensor parameters, $W$ and $\vec{b}$ are parameters of the linear part. $F$ is a non-linear function, and we use rectifier $F(z)=\max(0,z)$ in this paper.

The interaction can also be represented as a similarity score, such as cosine. We adopt neural tensor network here because it can capture more complicated interactions \cite{socher2013reasoning,socher2013recursive,DBLP:conf/ijcai/QiuH15}.

\subsection{Spatial RNN}
The second step of Match-SRNN is to apply a spatial RNN to the word level interaction tensor. Spatial RNN, also referred to as two dimensional RNN (2D-RNN), is a special case of multi-dimensional RNN \cite{DBLP:journals/corr/abs-0705-2011,graves2009offline,theis2015generative}. According to spatial RNN, given the representations of interactions between prefixes $S_1[1{:}i{-}1]\!{\sim}\! S_2[1{:}j]$, $S_1[1{:}i]\!{\sim}\! S_2[1{:}j{-}1]$ and $S_1[1{:}i{-}1]\!{\sim}\! S_2[1{:}j{-}1]$, denoted as $\vec{h}_{i-1,j},\vec{h}_{i,j-1}$, and $\vec{h}_{i-1,j-1}$, respectively, the interaction between prefixes $S_1[1{:}i]$ and $S_2[1{:}j]$ can be represented as follows:
\begin{equation}
\vec{h}_{ij}=f(\vec{h}_{i-1,j},\vec{h}_{i,j-1},\vec{h}_{i-1,j-1},\vec{s}_{ij}).
\end{equation}
Therefore we can see that spatial RNN can naturally model the recursive matching structure defined in Equation~(\ref{eq:RecursiveMatchStruct}).

For function $f$, we have different choices. The basic RNN usually uses a non-linear full connection layer as $f$. This type of function is easy for computing while often suffers from the gradient vanishing and exploding problem~\cite{DBLP:conf/icml/PascanuMB13}. Therefore, many variants of RNN has been proposed, such as Long Short Term Memory (LSTM) \cite{Hochreiter:1997:LSM:1246443.1246450}, Gated Recurrent Units (GRU)~\cite{cho2014learning} and Grid LSTM ~\cite{DBLP:journals/corr/KalchbrennerDG15}. Here, we adopt GRU since it is easy to implement and has close relationship with LCS as discussed in the following sections.

\begin{figure}[t]
\centering
\includegraphics[width=0.45\textwidth]{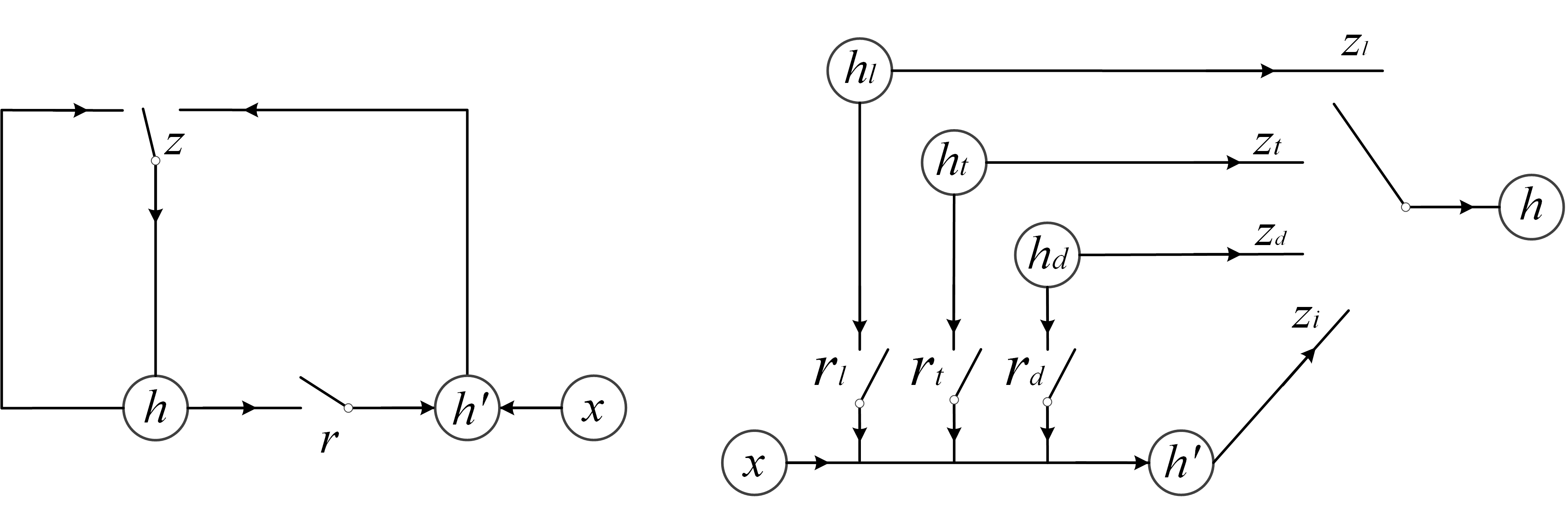}
\caption{Illustration of Gated Recurrent Units. The left one is 1D-GRU, where different $h$s are denoted as one node. The right one is the spatial GRU used in this paper. }
\label{fig:GRU}
\end{figure}

GRU is proposed to utilize several gates to tackle the aforementioned problems of basic RNN, and has shown excellent performance for tasks such as machine translation \cite{cho2014learning}. In this paper, we extend traditional GRU for sequences (1D-GRU) to spatial GRU. 
Figure~\ref{fig:GRU} describes clearly about the extensions.

For 1D-GRU , given a sentence $S{=}(x_1,x_2,\cdots,x_T)$, where $\vec{x}_t$ stands for the embedding of the $t$-th words, the representation of position $t$, i.e.~$\vec{h}_t$, can be computed as follows:
\begin{equation*}
	\begin{aligned}
		\vec{z} \,& {=} \sigma(W^{(z)}\vec{x}_t + U^{(z)}\vec{h}_{t-1}),\,\vec{r}{=} \sigma(W^{(r)}\vec{x}_t + U^{(r)}\vec{h}_{t-1}), \\
		\vec{h}'_t \,&{=} \phi(W\vec{x}_t{+}U(\vec{r}\odot \vec{h}_{t-1})),\,\vec{h}_t{=} (\vec{1}-\vec{z})\odot \vec{h}_{t-1} {+} \vec{z} \odot \vec{h}'_t,
	\end{aligned}
\end{equation*}
where $\vec{h}_{t-1}$ is the representation of position $t{-}1$, $W^{(z)},U^{(z)},W^{(r)},U^{(r)},W$ and $U$ are the parameters, $\vec{z}$ is the updating gate which tries to control whether to propagate the old information to the new states or to write the new generated information to the states, and $\vec{r}$ is the reset gate which tries to reset the information stored in the cells when generating new candidate hidden states.

When extending to spatial GRU, context information will come from three directions for a given position $(i,j)$, i.e.~$(i{-}1,j),(i,j{-}1)$ and $(i{-}1,j{-}1)$, therefore, we will have four updating gates $\vec{z}$, denoted as $\vec{z}_{l},\vec{z}_t, \vec{z}_d$ and $\vec{z}_i$, and three reset gates $\vec{r}$, denoted as $\vec{r}_l, \vec{r}_t, \vec{r}_d$. The function $f$ is computed as follows.
	\begin{align}\label{eq:SRNN}
&\vec{q}^T=[\vec{h}_{i-1,j}^T,\vec{h}_{i,j-1}^T,\vec{h}_{i-1,j-1}^T,\vec{s}_{ij}^T]^T,\nonumber\\
		&\vec{r}_l =\sigma(W^{(r_l)}\vec{q}+\vec{b}^{(r_l)}),\,\vec{r}_t =\sigma(W^{(r_t)}\vec{q}+\vec{b}^{(r_t)}),\nonumber\\
		&\vec{r}_d =\sigma(W^{(r_d)}\vec{q}+\vec{b}^{(r_d)}),\,\vec{r}^T  =[\vec{r}_l^T,\vec{r}_t^T,\vec{r}_d^T]^T,\nonumber\\
		&\vec{z}'_{i} = W^{(z_i)}\vec{q}+\vec{b}^{(z_i)},\,\vec{z}'_{l} =W^{(z_l)}\vec{q}+\vec{b}^{(z_l)},\nonumber\\
 		&\vec{z}'_{t} =W^{(z_t)}\vec{q}+\vec{b}^{(z_t)},\,\vec{z}'_{d}=W^{(z_d)}\vec{q}+\vec{b}^{(z_d)},\nonumber\\
 		&[\vec{z}_i, \vec{z}_l, \vec{z}_t, \vec{z}_{d}] = \texttt{SoftmaxByRow}([\vec{z}'_i, \vec{z}'_l, \vec{z}'_t, \vec{z}'_{d}]),\\
 		&\vec{h}'_{ij} {=}\phi(W\vec{s}_{ij} + U(\vec{r}\odot[\vec{h}_{i,j-1}^T, \vec{h}_{i-1,j}^T, \vec{h}_{i-1,j-1}^T]^T) + \vec{b}), \nonumber\\
 		&\vec{h}_{ij}     {=}\vec{z}_{l}\odot\vec{h}_{i,j-1}{+}\vec{z}_{t}\odot\vec{h}_{i-1,j}{+}
            		  \vec{z}_{d}\odot\vec{h}_{i-1,j-1}{+}\vec{z}_{i}\odot\vec{h}'_{ij},
	\end{align}
where $U$, $W$'s, and $b$'s are parameters, and $\texttt{SoftmaxByRow}$ is a function to conduct softmax on each dimension across the four gates, that is:
\begin{equation*}
[\vec{z}_p]_j=\frac{e^{[\vec{z'}_p]_j}}{e^{[\vec{z'}_i]_j}+ e^{[\vec{z'}_l]_j}+ e^{[\vec{z'}_t]_j}+ e^{[\vec{z'}_{d}]_j}}, \,\, p=i,l,t,d.
\end{equation*}

\subsection{Linear Scoring Function}
Since spatial RNN is a recursive model scanning the input from left top to right bottom, we can obtain the last representation as $\vec{h}_{mn}$ at the right bottom corner. $\vec{h}_{mn}$ reflects the global interaction between the two texts. The final matching score can be obtained with a linear function:
\begin{equation}
M(S_1,S_2)=W^{(s)}\vec{h}_{mn}+\vec{b}^{(s)},
\end{equation}
where $W^{(s)}$ and $\vec{b}^{(s)}$ denote the parameters.
\subsection{Optimization}
For different tasks, we need to utilize different loss functions to train our model.
Taking regression as an example, we can use square loss for optimization:
\begin{equation}
\mathcal{L}(S_1,S_2,y)= (y-M(S_1,S_2))^2,
\end{equation}
where $y\in R$ is the real-valued ground-truth label to indicate the matching degree between $S_1$ and $S_2$.

For ranking problem, we can utilize pairwise ranking loss such as hinge loss for training. Given a triple $(S_1,S_2^+,S_2^-)$, where the matching degree of $(S_1,S_2^+)$ is higher than $(S_1,S_2^-)$, the loss function is defined as:
\begin{equation*}
\mathcal{L}(S_1, S_2^+, S_2^-) = \max(0, 1-M(S_1, S_2^+)+M(S_1, S_2^-))
\end{equation*}
where $M(S_1,S_2^+)$ and $M(S_1,S_2^-)$ are the corresponding matching scores.

All parameters of the model, including the parameters of word embedding, neural tensor network, spatial RNN are jointly trained by BackPropagation and Stochastic Gradient Descent. Specifically, we use AdaGrad~\cite{duchi2011adaptive} on all parameters in the training process.

\section{Discussion}
In this section, we show the relationship between Match-SRNN and the well known longest common subsequence (LCS) problem.
\subsection{Theoretical Analysis}
The goal of LCS problem is to find the longest subsequence common to all sequences in a set of sequences (often just two sequences). In many applications such as DNA detection, the lengths of LCS are used to define the matching degree between two sequences. 

Formally, given two sequences, e.g.~$S_1{=}\{x_1,\cdots,x_m\}$ and $S_2{=}\{y_1,\cdots,y_n\}$, let $c[i,j]$ represents the length of LCS between $S_1[1{:}i]$ and $S_2[1{:}j]$. The length of LCS between $S_1$ and $S_2$ can be obtained by the following recursive progress, with each step $c[i,j]$ determined by four factors, i.e.~$c[i{-}1,j{-}1],c[i{-}1,j],c[i,j{-}1]$, and the matching between $x_i$ and $y_j$.
\begin{equation}\label{eq:LCS}
c[i,j]{=}\max(c[i,j{-}1], c[i{-}1,j],c[i{-}1,j{-}1]+\mathbb{I}_{\{x_i=y_j\}}),
\end{equation}
where $\mathbb{I}_{\{x_i=y_j\}}$ is an indicator function, it is equal to 1 if $x_i=y_j$, and $0$ otherwise. $c[i,j]{=}0$ if $i{=}0$ or $j{=}0$.

Match-SRNN has strong connection to LCS. To show this, we first degenerate the Match-SRNN to model an exact matching problem, by replacing the neural tensor network with a simple indicator function which returns 1 if the two words are identical and 0 otherwise, i.e.~$s_{ij}{=}\mathbb{I}_{\{x_i=y_j\}}$.
The dimension of spatial GRU cells is also set to 1. The reset gates of spatial GRU are disabled since the length of LCS is accumulated depending on all the past histories. Thus, Equation (4) can be degenerated as
\[
{h}_{ij}     ={z}_{l}\cdot{h}_{i,j-1}+{z}_{t}\cdot{h}_{i-1,j}+
            		  {z}_{d}\cdot{h}_{i-1,j-1}+{z}_{i}\cdot{h}'_{ij},
\]
where ${z}_{l}\cdot{h}_{i,j-1},  {z}_{t}\cdot{h}_{i-1,j}$, and ${z}_{d}\cdot{h}_{i-1,j-1}+{z}_{i}\cdot{h}'_{ij}$ correspond to the terms $c[i,j{-}1], c[i{-}1,j]$, and $c[i{-}1,j{-}1]+\mathbb{I}_{\{x_i=y_j\}}$ in Equation (\ref{eq:LCS}), respectively. Please note that $z_l, z_t, z_d$ and $z_i$ are calculated by \texttt{SoftmaxByRow}, and thus can approximate the $\max$ operation in Equation (\ref{eq:LCS}). By appropriately setting $z_i$ and $z_d$ and other parameters of Match-SRNN, $z_{d}\cdot h_{i-1,j-1}+z_{i}\cdot h'_{ij}$ can approximate the simple addition operation ${h}_{i-1,j-1}+{s}_{ij}$, where ${h}_{i-1,j-1}$ and ${s}_{ij}$ correspond to the $c[i{-}1,j{-}1]$ and $\mathbb{I}_{\{x_i=y_j\}}$, respectively.
Therefore, the computation of ${h}_{ij}$ in Eq.~(4) can well approximate $c[i,j]$ in Eq.~(\ref{eq:LCS}).

\subsection{Simulation Results}
\begin{figure}[t]
\centering
\includegraphics[width=0.44\textwidth]{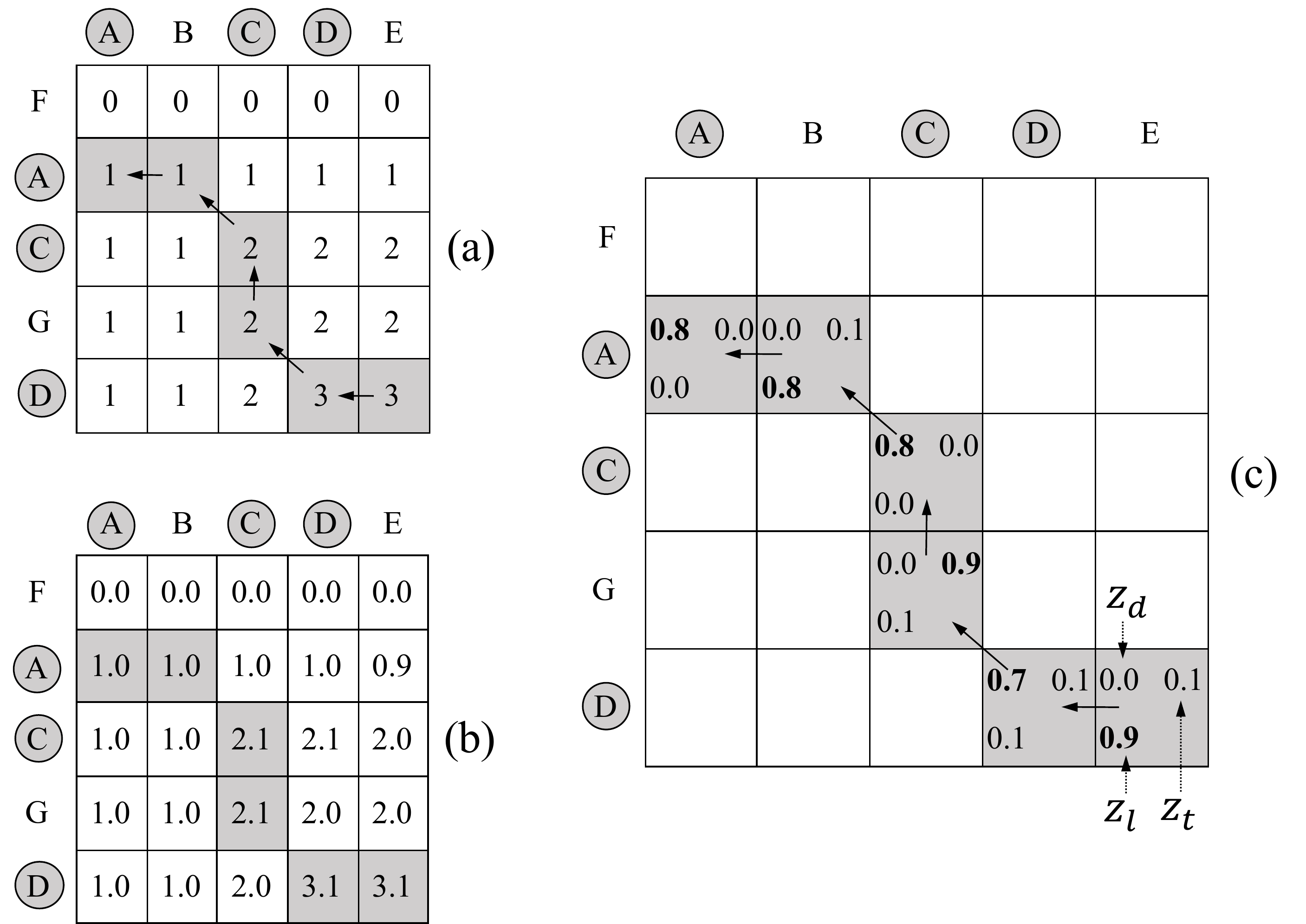}
\caption{The simulation result of LCS by Match-SRNN. Figure (a) shows the matching degree and path discovered in LCS. Figure (b) shows the simulation results of Match-SRNN. Figure (c) shows the backtracing process of finding the gray path in Match-SRNN.}\label{fig:simulation}
\end{figure}

We conducted a simulation experiment to verify the analysis result shown above. The dataset was constructed by many random sampled sequence pairs, with each sequence composed of characters sampled from the vocabulary \textit{\{A B C D E F G H I J\}}. Firstly, the dynamic programming algorithm of LCS was conducted on each sequence pair, and the normalized length of LCS is set to be the matching degree of each sequence pair. For simulation, we split the data into the training (10000 pairs) and testing set (1000 pairs), and trained Match-SRNN with regression loss. The simulation results on two sequences $S_1=\{A,B,C,D,E\}$ and $S_2=\{F,A,C,G,D\}$ are shown in Figure~\ref{fig:simulation}.

Figure~\ref{fig:simulation} (a) shows the results of LCS, where the scores at each position $(i,j)$ stands for $c[i,j]$, and the gray path indicates the process of finding the LCS between two sequences, which is obtained by backtracing the dynamic programming process. Figure~\ref{fig:simulation} (b) gives the results of Match-SRNN, where the score at each position $(i,j)$ stands for the representation $\vec{h}_{ij}$ (please note that for simplification the dimension of $\vec{h}_{ij}$ is set to $1$). We can see that the scores produced by Match-SRNN is identical to that obtained by LCS, which reveals the relationship between Match-SRNN and LCS.

The gray path in Figure~\ref{fig:simulation} (b) shows the main path of how local interactions are composited to the global interaction, which
is generated by backtracing the gates. Figure~\ref{fig:simulation} (c) shows the path generation process, where the three values at each positions stands for the three gates, e.g.~$z_l{=}0.9,z_t{=}0.1,z_d{=}0$ at position $(5,5)$. Considering the last position $(5,5)$, the matching signals are passed over from the direction with the largest value of gates, i.e.~$z_l$, therefore, we move to the position $(5,4)$. At position $(5,4)$, the largest value of gates is $z_d{=}0.7$, therefore, we should move to position $(3,3)$. We can see that the path induced by Match-SRNN is identical to that of by dynamic programming. This analysis gives a clear explanation on the mechanism of how the semantic matching problem be addressed by Match-SRNN.

\section{Experiments}
We conducted experiments on the tasks of question answering (QA) and paper citation (PC) to evaluate the effectiveness of Match-SRNN.

QA dataset is collected from Yahoo! Answers, a community question answering system where some users propose questions to the system and other users will submit their answers, as in~\cite{wan2016aaai}.
The whole dataset contains 142,627 (question, answer) pairs, where each question is accompanied by its best answer.
We select the pairs in which questions and their best answers both have a length between 5 and 50.
After that the dataset contains 60,564 (questions, answer) pairs which form the positive pairs. For each question, we first use its best answer as a query to retrieval the top 1,000 results from the whole answer set, with Lucene search engine. Then we randomly select 4 answers from them to construct the negative pairs. 

PC task is to match two papers with citation relationship.
The dataset is constructed as in ~\cite{pang2016aaai}.
The paper abstract information and citation network are collected from a commercial academic website.
The negative pairs are randomly sampled from the whole dataset.
Finally, we have 280K positive and 560K negative instances.

\subsection{Effectiveness of Match-SRNN}
We compared Match-SRNN with several existing deep learning methods, including ARC-I, ARC-II, CNTN, LSTM-RNN, MultiGranCNN, MV-LSTM and MatchPyramid. We also compared with BM25~\cite{robertson1995okapi}, which is a popular and strong baseline for semantic matching in information retrieval. For Match-SRNN, we also implemented the bidirectional version for comparison, which also scans from right bottom to left top on the word interaction tensor, denoted as Bi-Match-SRNN.

In our experiments, we set the parameters and the baselines as follows.
Word embeddings used in our model and in some baseline deep models are all initialized by \texttt{SkipGram} of \texttt{Word2Vec} \cite{mikolov2013distributed}.
Following the previous practice, word embeddings are trained on the whole question answering data set, and the dimension is set to 50.
The batch size of SGD is set to 128. All other trainable parameters are initialized randomly by uniform distribution with the same scale, which is selected according to the performance on validation set. The initial learning rates of AdaGrad are also selected by validation. The dimension of neural tensor network and spatial RNN is set to 10, because it won the best validation results among the settings of $d=1,2,5,10,$ and $20$. The other parameters for the baseline methods are set by taking the values from the original papers.

The QA task is formulated as a ranking problem. Therefore, we use the hinge loss for optimization, as shown in Section 4.4, and the results are evaluated by typical ranking measures, such as Precision at 1 (denoted as P@1) and Mean Reciprocal Rank (MRR).
\begin{equation*}
P@1=\frac{1}{N}\sum_{i=1}^{N}\delta(r(S_2^{+(i)})= 1),\,
MRR=\frac{1}{N}\sum_{i=1}^{N}\frac{1}{r({S_{2}^{+(i)}})},
\end{equation*}
where $N$ is the number of testing ranking lists, $S_2^{+(i)}$ is the positive sentence in the $i-th$ ranking list, $r(\cdot)$ denotes the rank of a sentence in the ranking list, and $\delta$ is the indicator function.
The PC task is formulated as a binary classification task. Therefore the matching score is used by a softmax layer and cross entropy loss is used for training.
We use classification accuracy (Acc) as the evaluation measure.

\begin{table}[t]
\begin{tabular}{l|cc|c} \hline
&
\multicolumn{2}{c|}{QA} &
\multicolumn{1}{c}{PC} \\ \hline
\multicolumn{1}{c|}{Model} &
\multicolumn{1}{c}{P@1} &
\multicolumn{1}{c|}{MRR} &
\multicolumn{1}{c}{Acc} \\ \hline
Random Guess			& 0.200 & 0.457 & 0.500   \\
BM25        			& 0.579 & 0.726 & 0.832	    \\ \hline
ARC-I 					& 0.581 & 0.756 	& 0.845	\\
CNTN					& 0.626 & 0.781 	& 0.862 	\\
LSTM-RNN   		  		& 0.690 	& 0.822 	& 0.878 	\\ \hline
MultiGranCNN			& 0.725 	& 0.840 & 0.885		\\
MV-LSTM  				& 0.766	& 0.869	& 0.890	\\ \hline
ARC-II 					& 0.591 & 0.765 	& 0.865	\\
MatchPyramid-Tensor			& 0.764 & 0.867 	& 0.894\\ \hline
Match-SRNN 				& 0.785 & 0.879 & 0.898 \\
Bi-Match-SRNN  			& \textbf{0.790}	& \textbf{0.882} 	& \textbf{0.901}	\\
\hline
\end{tabular}
\centering
\caption{Experimental results of QA and PC tasks.}\label{tab:results}
\end{table}
The experimental results are listed in Table~\ref{tab:results}. We have the following experimental findings:

(1) By modeling the recursive matching structure, Match-SRNN can significantly improve the performances, compared with all of the baselines.
Taking QA task as an example, compared with BM25, the improvement is about $36.4\%$ in terms of P$@$1. Compared with MV-LSTM, the best one among deep learning methods focusing on learning sentence representations, the improvement is about $3.1\%$. Compared with the deep models using hierarchical composition structures (i.e.~ARC-II and MatchPyramid), the improvements are at least $3.4\%$. For PC task, Match-SRNN also achieves the best results, though the improvements are smaller as compared to those on QA task. This is because the task is much easier, and even simple model such as BM 25 can produce a good result. From the above analysis, we can see that the recursive matching structure can help to improve the results of semantic matching.

(2) Both of the two matching paradigms (representing text into dense vectors and modeling the interaction relationship) have their own advantages, and the results are comparable, e.g.~the previous best results of the two paradigms on QA dataset are 0.766/0.869 (MV-LSTM) and 0.764/0.867 (MatchPyramid).


\subsection{Visualization}
To show how Math-SRNN works and give an insight on its mechanism on real dataset, we conducted a case study to visualize the interactions generated by Match-SRNN.

The example sentences are selected from the testing set of QA dataset.

{\em Question: ``How to get rid of \textbf{memory stick error} of my sony cyber shot?}"

{\em Answer: ``You might want to try to format the \textbf{ memory stick} but what is the \textbf{error} message you are receiving.}"

We can see that in this example, the matching of a bigram ($memory$, $stick$) and a keyword ($error$) is important for calculating the matching score. In this experiment, we used a simplified version Match-SRNN to give a better interpretation. Specifically, we set the values of different dimensions in the gates to be identical, which is convenient for the backtracing process. Since the hidden dimension is set to 10, as used in the above Match-SRNN, we can obtain 10 values for each $\vec{h}_{ij}$. We choose to visualize the feature map of the dimension with the largest weight in producing the final matching score. Similar visualization can be obtained for other dimensions, and we omit them due to space limitation.

\begin{figure}[t]
\includegraphics[width=0.4\textwidth]{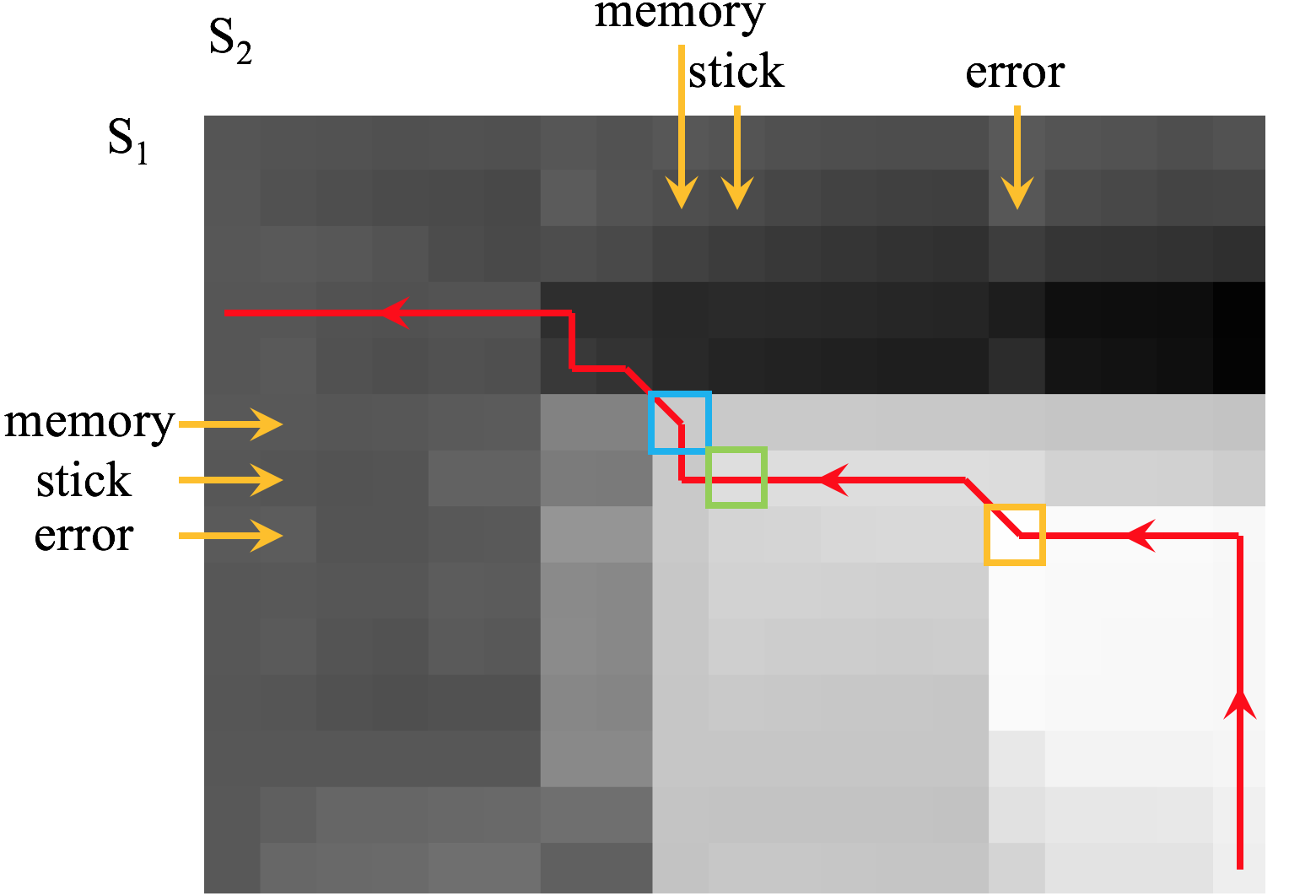}
\caption{A representative interaction learned by Match-SRNN, where the brightness is dependent on the interaction value at each position, and the path in red denotes the information diffussion process generated by backtracing the maximum gates.}
\label{fig:tensor}
\end{figure}

The visualization results are shown in Figure~\ref{fig:tensor}, where the brightness of each position stands for the interaction strength. We can see that the recursive matching structure can be shown clearly. When there is a strong word level interaction happened in some position (e.g., the exact word match of $(memory, memory)$), the interaction between the two texts are strengthened and thus the bottom-right side of the position becomes brighter. The interactions are further strengthened with more strong word level interactions, i.e., the bottom-right side of the matching positions of $(stick,stick)$ and $(error, error)$ become even brighter. Backtracing the gates, we obtain the matching path which crosses all the points with strong word interactions, as shown by red curve in Figure~\ref{fig:tensor}. It gives a clear interpretation on how Match-SRNN conducted the semantic matching on this real example.
\section{Conclusions}
In this paper, we propose a recursive thinking to tackle the complicated semantic matching problem.
Specifically, a novel deep learning architecture, namely Match-SRNN is proposed to model the recursive matching structure. Match-SRNN consists of three parts: a neural tensor network to obtain the word level interactions, a spatial RNN to generate the global interactions recursively, and a linear scoring function to output the matching degree. Our analysis reveals an interesting connection of Match-SRNN to LCS. Finally, our experiments on semantic matching tasks showed that Match-SRNN can significantly outperform existing deep learning methods. Furthermore, we visualized the recursive matching structure discovered by Match-SRNN on a real example.

\newpage
\section*{Acknowledgments}
This work was funded by 973 Program of China under Grants No. 2014CB340401 and 2012CB316303, 863 Program of China under Grants No. 2014AA015204, the National Natural Science Foundation of China (NSFC) under Grants No. 61232010,61472401, 61433014, 61425016, 61203298, and 61572473, Key Research Program of the Chinese Academy of Sciences under Grant No. KGZD-EW-T03-2, and Youth Innovation Promotion Association CAS under Grants No. 20144310 and 2016102.


\end{document}